\documentclass[10pt,twocolumn,letterpaper]{article}

\usepackage{cvpr}
\usepackage{times}
\usepackage{epsfig}
\usepackage{graphicx}
\usepackage{amsmath}
\usepackage{amssymb}
\usepackage{lipsum}
\usepackage{multirow}
\usepackage{array}
\usepackage{float}

\usepackage[pagebackref=true,breaklinks=true,letterpaper=true,colorlinks,bookmarks=false]{hyperref}

\cvprfinalcopy 

\def\cvprPaperID{107} 

\ifcvprfinal\fi
\begin{document}

\title{Structured Feature Learning for Pose Estimation}

\author{Xiao Chu \quad Wanli Ouyang \quad Hongsheng Li \quad Xiaogang Wang\\
Department of Electronic Engineering, The Chinese University of Hong Kong\\
{\tt\small xchu@ee.cuhk.edu.hk} \quad
{\tt\small wlouyang@ee.cuhk.edu.hk} \quad
{\tt\small hsli@ee.cuhk.edu.hk} \quad
\\
{\tt\small xgwang@ee.cuhk.edu.hk}
}

\maketitle

\begin{abstract}
In this paper, we propose a structured feature learning framework to reason the correlations among body joints at the feature level in human pose estimation.
Different from existing approaches of modeling structures on score maps or predicted labels, feature maps preserve substantially richer descriptions of body joints. The relationships between feature maps of joints are captured with the introduced geometrical transform kernels, which can be easily implemented with a convolution layer. 
Features and their relationships are jointly learned in an end-to-end learning system. A bi-directional tree structured model is proposed, so that the feature channels at a body joint can well receive information from other joints. The proposed framework improves feature learning substantially. With very simple post processing, it reaches the best mean PCP on the LSP and FLIC datasets. Compared with the baseline of learning features at each joint separately with ConvNet, the mean PCP has been improved by $18\%$ on FLIC. The code is released to the public.
\footnote{The code can be found at 
\url{http://www.ee.cuhk.edu.hk/~xgwang/projectpage_structured_feature_pose.html}. For more technical details, please contact the corresponding authors Wanli Ouyang and Xiaogang Wang}
\end{abstract}

\section{Introduction}
Human pose estimation is to estimate the locations of body joints from images.
It can assist a variety of vision tasks such as action recognition \cite{wang2013approach,xu2012combining,liang2014expressive}, tracking \cite{cho2013adaptive}, person re-identification \cite{xiao2016learning}, and human computer interaction.
Despite the long history of efforts, it is still a challenging problem.
The large variation in limb orientation, clothing, viewpoints, background clutters, truncation, and occlusion make localization of body joints difficult.

\begin{figure}[t]
\begin{center}
   \includegraphics[width=1\linewidth]{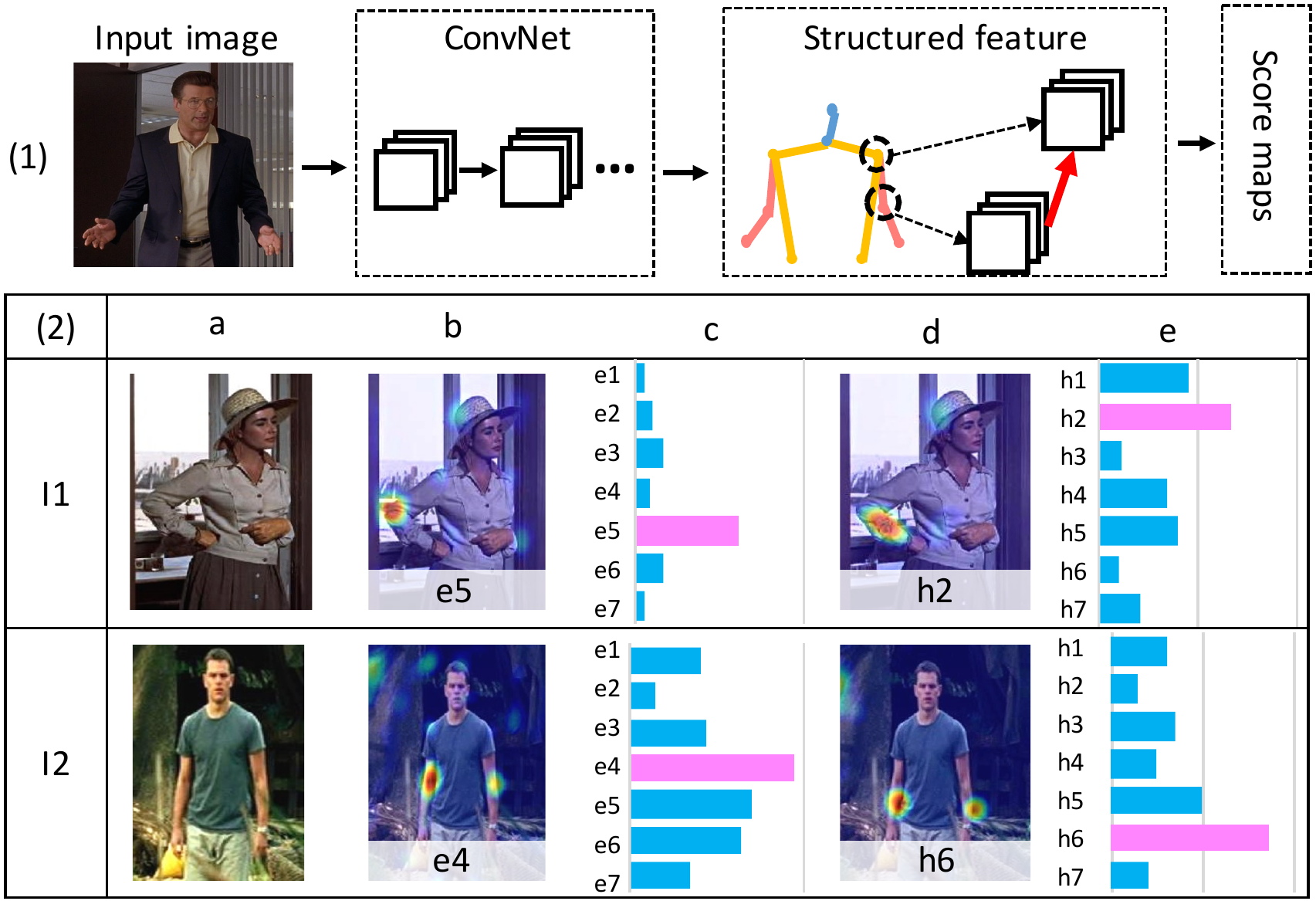}
\end{center}
   \caption{ (1) Our approach jointly learns  feature maps at different body joints and the spatial and co-occurrence relationships between feature maps.
	The information from different joints passes at the feature level. (a) Two input images (I1 and I2) with different poses. (c) Responses of feature channels for elbow ($e1$-$e7$). (I1, b) is the response map of $e5$ for image I1. (I2, b) is the response map of $e4$ for image I2. Similarly, (d) and (e) show the response maps and responses of different feature channels for lower arm.}
\label{fig:firstpage}
\end{figure}

Independent prediction of body joint locations from appearance score maps can be refined by modeling the spatial relationship among correlated body joints \cite{yang2013articulated,chen2014articulated,pishchulin2013poselet}.
On score maps, the information at a location is summarized into a single probability value, indicating the likelihood of the existence of the corresponding body joint.
For example, if a location on the score map of elbow has a large response, we can only reach the conclusion that this location may belong to elbow, but cannot tell the in-plane and out-plane rotation of the elbow, the orientations of the upper arm and the lower arm associated with it, whether it is covered with clothes, and its occlusion status. Such detailed information is valuable for predicting the locations of other body joints, but is missed from the score maps, which makes structural learning among body joints much less effective. 

We observe that these types of information are well preserved at the feature level, where hierarchical feature representations are learned with Convolutional Networks (ConvNets) \cite{Krizhevsky:ImageNetCNN,zeiler2013visualizing,sermanet2013overfeat,simonyan2014very,szegedy2014going}. Fig. \ref{fig:firstpage} shows the responses of feature maps of elbow and lower arm for different input images. Given the V-shaped elbow covered with clothes in I1, the feature channel $e5$ has the largest response as shown in (I1, c). 
In the meanwhile, the feature channel $h2$ for lower arm has the largest response in (I1, e).
Given the straight elbow uncovered with clothes in I2, the feature channels $e4$ and $h6$ have the largest responses to elbow and lower arm respectively. 
It indicates that different feature channels are activated for different visual patterns. The feature maps of different joints also have strong correlations. In Fig. \ref{fig:firstpage}, $e5$ is positively correlated with $h2$ and anti-correlated with $h6$. 
Both the spatial distribution of the responses and the semantic meaningful description of body joints are encoded at the feature maps by activating different channels.
 
Some existing works \cite{yang2013articulated,chen2014articulated,pishchulin2013poselet} employed mixtures clustered from spatial configuration among neighboring body joints. However, the number of mixtures for each body joint (fewer than 20) is incomparable to hundreds of feature channels from ConvNets, which not only include spatial configuration of body joints, but also other information such as occlusion status and clothing. 
Hence, we propose to exploit the structure information of body joints at the feature level.
Our proposed approach shows that the spatial and co-occurrence relationship among feature maps can be modeled by a set of geometrical transform kernels. These kernels can be implemented with convolution and the relationships can be learned in and end-to-end learning system.

It is important to design proper information flow between body joints, so that features at a joint can be optimized by receiving messages from highly correlated joints and will not be disturbed by less correlated joints in distance. A bi-directional tree-structured model is proposed. The proposed model connects correlated joints and passes messages in both directions along the tree. Therefore, every joint can receive information from all the neighboring joints. 


The contributions of this work are summarized as three-fold. First, it proposes an end-to-end learning framework to capture rich structural information among body joints at the feature level for pose estimation. Second, it is shown that the relationships among feature maps of neighboring body joints can be learned by the introduced geometrical transform kernels and can be easily implemented with convolutional layers. Third, a bi-directional tree-structured model is proposed, so that each joint can receive information from all the correlated joints and optimize its features. 

Experimental results show that the proposed approach can improve feature learning substantially. Compared with learning features at each joint separately with ConvNet, it improves the mean PCP by $18\%$ on the FLIC dataset. It also reaches the highest mean PCP $80.8\%$ on the LSP dataset and $95.2\%$ on the FLIC dataset. This work focuses on feature learning and only adopts very simple post processing. 
It already outperforms the state-of-the-art method which employed sophisticated post processing techniques with a large margin. 


\section{Related Works}
Previous pose estimation works can be divided into two groups.
The first is to model the geometrical distribution of body joints \cite{yang2013articulated,wang2013beyond,Wang:PoseletsParsing,pishchulin2013poselet,eichner2012appearance,cho2013adaptive,pishchulin14CVPR,felzenszwalb2005pictorial,tian2012exploring,Andriluka:Pictorial,sapp2013modec,ouyang2014multi,yang2016end} which can be viewed as post processing on detection score maps and prediction labels. They are mainly based on hand-crafted features.
The Pictorial Structure Model \cite{felzenszwalb2005pictorial} defined pairwise terms to represent relationship between body joint locations.
Later, Yang \etal \cite{yang2013articulated} proposed the flexible mixture-of-parts model to combine part detection results with a tree-structured model, which provided simple and exact inference.
Nevertheless, it is believed that the tree-structured model is ``oversimplified''.
In light of this, many works introduced more complex structures, and researchers have obtained improvement in performance.
Loopy structure \cite{Wang:PoseletsParsing}, latent variable \cite{tian2012exploring}, poselet \cite{pishchulin2013poselet, Wang:PoseletsParsing} and strong appearance \cite{pishchulin13iccv} modeled structural information at different levels.
They investigated different structures to model the spatial constraints among body joints on score maps. 
In our work, a bi-directional tree is used to model the correlation among feature maps. 
In the future, the investigations on structures in previous works can be incorporated in our framework to guide the message passing at the feature level.

The second group focus on more powerful feature generators such as ConvNets \cite{toshev2014deeppose,fan2015combining,chen2014articulated,tompson2014joint,tompson2015efficient,carreira2015human,fan2015combining}.
The use of deep models brings large progress \cite{kang2016object,Ouyang2015DeepID}.
DeepPose \cite{toshev2014deeppose} used ConvNet to regress joint locations with multiple steps.
Chen \etal \cite{chen2014articulated} used ConvNet features and built up image-dependent pairwise relations to measure relationship among body joints.
Fan \etal \cite{fan2015combining} combined local and global features to jointly predict joint locations.
Tompson \etal \cite{tompson2014joint,tompson2015efficient} implemented the multi-resolution deep model and Markov random field within an end-to-end joint training framework.
Carreira and Malik \cite{carreira2015human} proposed to build up dependency among input and output spaces. 
In order to iteratively refine prediction results, they concatenated the body joint location predictions at the previous steps with the image as the input of current step.
However, existing ConvNet models either learned the pair-wise relationship among body joints from score maps or did not learn pair-wise relationship. 
Learning relationship among parts at the feature level was not investigated.


\begin{figure}[ht]
\begin{center}
   \includegraphics[width=1\linewidth]{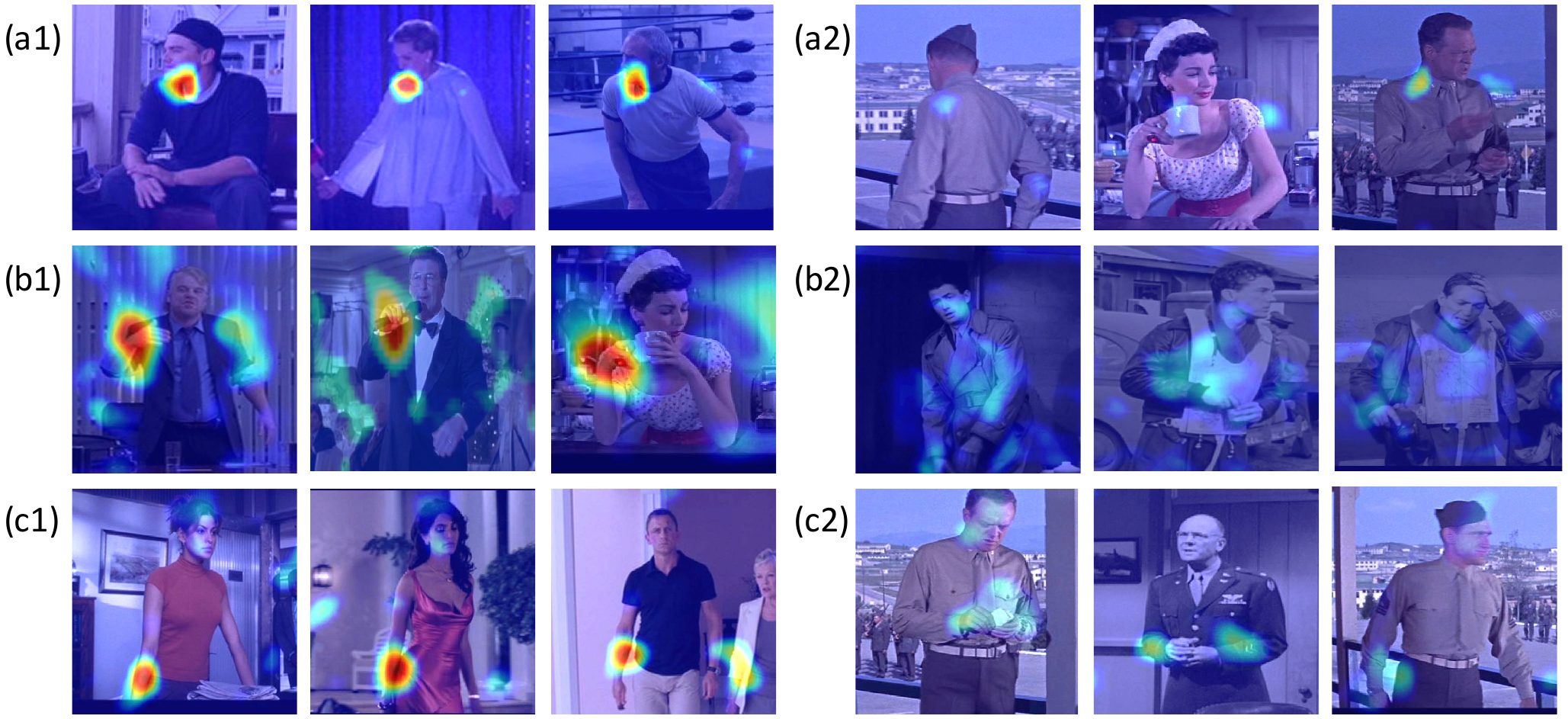}
\end{center}
\vspace{-2 mm}
   \caption{Examples of response maps of different images to the same feature channels. (a) A feature channel for the neck. (b) A feature channel for the left wrist. (c) A feature channel for the left lower arm.}
\label{fig:TopK}
\end{figure}

\section{Structural Feature Learning}
\subsection{Feature maps of body joints}
\label{SEC: feature message}
ConvNets employ multiple layers to learn hierarchical feature representations of input images.
Features in lower layers capture low-level information, while those in higher layers can represent more abstract concepts, such as poses, attributes and object categories. Widely used ConvNets (e.g. AlexNet \cite{Krizhevsky:ImageNetCNN}, Clarifai \cite{zeiler2013visualizing}, Overfeat \cite{sermanet2013overfeat}, GoogleNet \cite{szegedy2014going}, and VGG \cite{simonyan2014very}) employ fully connected (fc) layers following convolutional layers to capture the global information. 
In fully convolutional nets (fcn), ${1}\times{1}$ convolution is used to replace fc layers. 
In this work, we use fully convolutional VGG net \cite{simonyan2014very} as the base model and extract feature maps in the fcn7 layer. 

Each body joint has a separate set of $128$ feature maps. All the joints share lower layers up to the fcn6 layer, which has $4,096$ feature channels. Denote $\mathbf{h}_{fcn6}(x,y)$ as the feature vector obtained at location $(x,y)$ in the fcn6 layer and it is a $4,096$ dimensional vector. The $128$ dimensional feature vector for body joint $k$ at $(x,y)$ in the fcn7 layer is computed as 
\begin{equation}
   \mathbf{h}_{fcn7}^k (x,y) = f(\mathbf{h}_{fcn6}(x,y) \otimes {\mathbf{w}^k_{fcn7}}+\mathbf{b}_{fcn6}),
   \label{eq:firsteq}
\end{equation}
where $\otimes$ denotes convolution, $f$ is a nonlinear function, $\mathbf{w}^k_{fcn7}$ is the filter bank for joint $k$ including $128$ filters, $\mathbf{b}_{fcn6}$ is the bias, and $\mathbf{h}_{fcn7}^k$ is the feature tensor contains $128$ feature maps for joint $k$.

The feature maps of body joints contain rich information and detailed descriptions of human poses and appearance. Fig. \ref{fig:TopK} shows the response maps of different images to the same feature channels. In (a1) and (a2), a feature channel for the neck is chosen. All the images in (a1) have high responses to this feature channel and the highest responding regions locate on necks. Persons in these images all look to the left with similar 3D orientations of head. 
Images in (a2) have much lower responses to this feature channel and their highest responding regions distribute randomly. Persons in these images have various head orientations different than those in (a1). Therefore, this feature channel captures specific head orientations. Similarly, the feature channel for the left wrist in (b) describes left wrists occluding left shoulders when persons hold cups or cell phones. 
The feature channel in (c) can effectively localize downward lower arms without clothes covered. 

\begin{figure*}[t]
\begin{center}
\includegraphics[width=1\linewidth]{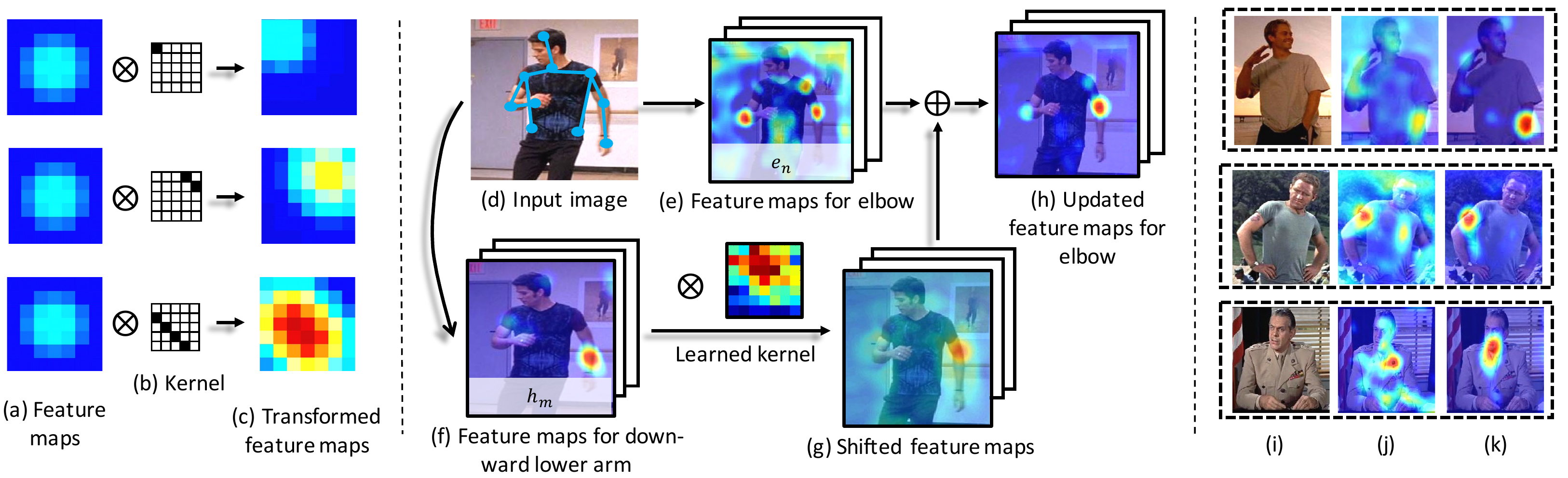}
\end{center}
\vspace{-2 mm}
   \caption{(a)-(c) show that feature maps can be shifted through convolution with kernels. 
   (d)-(h) show an example of updating feature maps by passing information between joints.
   (i)-(k) compare the featuer maps before (\ie (j)) and after (\ie (k)) information passing.}
\label{fig:geometric feature transfer}
\label{fig:change of receptive field}
\end{figure*}

\subsection{Information passing}
\label{Sec: Structural feature learning}

Since spatial distributions and semantic meaning of feature maps obtained at different joints are highly correlated, passing the rich information contained in feature maps between joints can effectively improve features learned at each joint. 
In previous works, messages could be passed by distance transfer \cite{girshick2014deformable,yang2013articulated,Ouyang2015DeepID} and Conditional Random Field (CRF) \cite{zheng2015conditional,krahenbuhl2012efficient}. 
We show that under a fully convolutional neural network, messages can be passed between feature maps through the introduced geometrical transform kernels. The FCN filters and the kernels can be jointly learned. 

Fig. \ref{fig:geometric feature transfer} (a)-(c) shows that convolution with asymmetric kernels could geometrically shift the feature responses. 
(a) is a feature map assuming Gaussian distribution. (b) are different kernels for illustration. (c) are the transformed feature maps after convolution. 
The feature map has been shifted towards different directions and sum up to different values.


In order to illustrate the process of information passing, an example is shown in Figure \ref{fig:geometric feature transfer} (d)-(g). Given an input image in (d), its feature maps for elbow and lower arm are shown in (e) and (f). One of the lower-arm feature maps $h_m$ has high response, since its feature channel describes downward lower arm without clothes covered. Another elbow feature map $e_n$ also has high response and it is positively correlated with $h_m$. 
One expects to use $h_m$ to reduce false alarms and enhance the responses on the right elbow. It is not suitable to directly add $e_n$ to $h_m$, since there is a spatial mismatch between the two joints. Instead, we first shift $h_m$ towards the right elbow through the geometrical transform kernels and then add the transformed feature maps to $e_n$. The refined feature maps in (h) have much better prediction. Since each feature map captures detailed pose information of the joint, the relative spatial distribution between the two maps is stable and the kernel can be easily learned. 
Since some elbow feature maps may be anti-correlated with $h_m$, their kernels could have negative values to prevent unrelated feature channels from generating false alarms. (i)-(k) show more examples to demonstrate the effectiveness of information passing between joints on feature learning. The geometric constraints among body joints could be consolidated by shifting feature map of one body joint towards its nearby joints. 
The information passing described above can be easily implemented with convolution layers.

\subsubsection{Stacked transform kernels}
The kernel size decides how far a feature map can be shifted. In order to reduce the number of parameters and also support the cases when neighboring joints are in  distance, we employ successive convolutions geometrical transform kernels to approximate a large kernel.
Each convolution is followed by a nonlinear transform. 
In our approach, the neighbor joints are defined with a tree structure as shown in Fig. \ref{fig:outline}. According to the statistics on our datasets, the largest distance between neighbor joints is within $72$ pixels on FLIC dataset, such target joint can be reached by three successive $7\times7$ geometrical transform kernels. 

\subsubsection{Bi-directional tree}
\label{sec:tree}
\begin{figure*}[t!]
\begin{center}
   \includegraphics[width=0.9\linewidth]{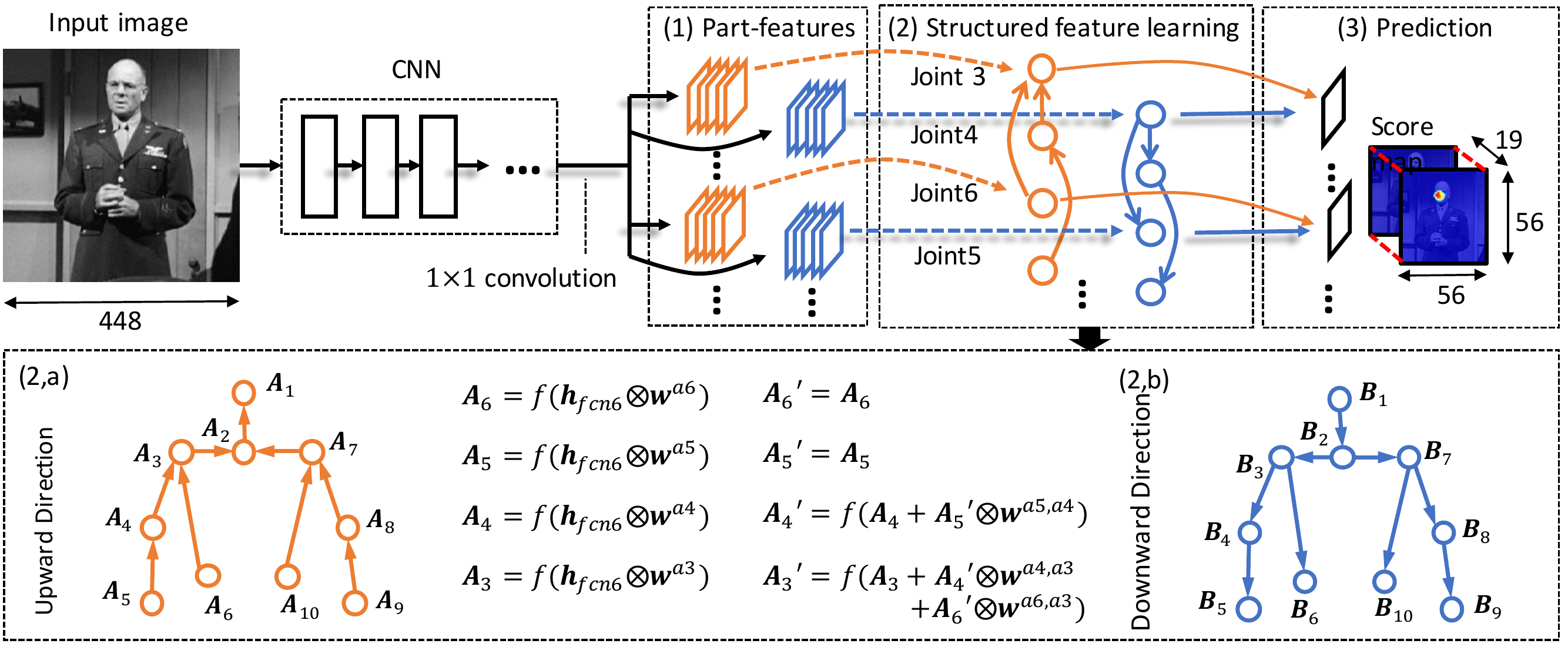}
\end{center}
\vspace{-2 mm}
   \caption{Our pipeline for pose estimation. (1) Original feature maps for body joints. (2) Refine the feature maps by information passing in a structure feature learning layer. (2,a) and (2,b) show the details of the bi-directional tree which have information flows in opposite directions. 
   The process of updating feature maps are also illustrated. (3) Predict score maps for joints based on feature maps. Dashed line is copy operation and solid line is convolution.}
\label{fig:outline}
\end{figure*}

To optimize features obtained at a joint, one expect to receive information from all the other joints with a fully connected graph. 
It has two drawbacks.
First, in order to directly model the relationship between feature maps of joints in distance, large transform kernels have to be introduced and they are difficult to learn. 
Second, the relationship between some joints (such as head and foot) are unstable. 
A better way is to propagate information between them through intermediate joints on a designed graph.
The neighbor joints on the graph are close in distance and have relatively stable relationship in the graph. In this work, a tree structure shown in Fig. \ref{fig:outline} (2,a) and (2,b) is chosen. 

In Fig. \ref{fig:outline} (2,a), information flows from leaf joints to root joints. Let $\{\mathbf{A}_k\}$ be the original feature maps directly obtained from the fcn6 layer.
Here, $\{\mathbf{A}_k\}$ is the concrete case of $\mathbf{h}^k_{fcn6}$ in Eq. \ref{eq:firsteq}.
$k$ is the index of joint.
The refined feature maps after message passing are denoted by $\{\mathbf{A'}_k\}$.

\begin{equation}
	\mathbf{A}_k = f(\mathbf{h}_{fcn6} \otimes \mathbf{w}^{a_k}),
\end{equation}
where $\mathbf{h}_{fcn6}$ are the fcn6 feature maps, $\mathbf{w}^{a_k}$ is the filter bank for joint $k$, and $f$ is the rectified linear unit. The process of refining features is explained below. 

Since $\mathbf{A}_5$ and $\mathbf{A}_6$ are at the leaf joints in the upward direction tree, they do not receive information from other joints, so the refined feature maps are the same as the original ones, i.e. 

\begin{equation}
	\mathbf{A'}_5 = \mathbf{A}_5, \hspace{0.1in} \mathbf{A'}_6 = \mathbf{A}_6. 
\end{equation}
$\mathbf{A}_4$ is updated by receiving information from $\mathbf{A'}_5$,
\begin{equation}
	\mathbf{A'}_4 = f(\mathbf{A}_4 + \mathbf{A'}_5 \otimes \mathbf{w}^{a_5,a_4}), 
\end{equation}
where $\mathbf{w}^{a_5,a_4}$ is a collection of transform kernels between joint $5$ and joint $4$. $\mathbf{A}_3$ is updated by receiving information from both $\mathbf{A'}_4$ and $\mathbf{A'}_6$,
\begin{equation}
	\mathbf{A'}_3 = f(\mathbf{A}_3 + \mathbf{A'}_4 \otimes \mathbf{w}^{a_4,a_3} + \mathbf{A'}_6 \otimes \mathbf{w}^{a_6,a_3}). 
\end{equation}
Feature maps of other joints are updated in a similar way. 

To obtain complementary features, we design another branch with the same tree structure but opposite information flow in Fig. \ref{fig:outline} (2,b). The original feature maps $\{\mathbf{B}_k\}$ are obtained in the same way as $\{\mathbf{A}_k\}$, but the refined feature maps $\{\mathbf{B'}_k\}$ are updated in the opposite order as indicated by the arrows' direction in Fig. \ref{fig:outline} (2,b).
The final feature maps at each node are obtained by concatenating the two sets of updated feature maps $[\mathbf{A'}_k, \mathbf{B'}_k]$. The concatenated 256 channel feature tensor for joint $k$ is used to predict the score map of joint $k$ in a later step. 

\begin{table*}[ht]\scriptsize
\centering
\begin{tabular}{p{0.8cm}  | l  | p{1.1cm}<{\centering} | p{1.2cm}<{\centering} | p{1.2cm}<{\centering} | p{1.2cm}<{\centering} | p{1.2cm}<{\centering} | p{0.9cm}<{\centering} | p{1cm}<{\centering} | p{1cm}<{\centering} | p{1cm}<{\centering} }
\hline\hline
Name & Description & 1 & 2 & 3 & 4 & 5 & 6 & 7 & 8 & 9 \\
\hline
\multirow{2}{4em}{VGG-16} 
& filter & conv1\_1,2 &max &conv2\_1,2 &max
& conv3\_1,2,3 & max & conv4\_1,2,3 &max   & conv5\_1,2,3 \\
& channel(kenel-stride) 
& 64(3-1) &64(2-2) &128(3-1) & 128(2-2) & 256(3-1) &256(2-2)
                     &512(3-1)     & 512(2-2) & 512(3-1) \\
\hline
\multirow{2}{4em}{Ours}
& filter 
& conv1\_1,2 &max &conv2\_1,2 &max
& conv3\_1,2,3 & max & conv4\_1,2,3 & conv5\_1,2,3 & fcn6 \\
 & channel(kenel size) 
  & 64(3-1) &64(2-2) &128(3-1) & 128(2-2) & 256(3-1) &256(2-2)
                     &512(3-1)  & 512(3-1) &4096(7-1) \\ 
\hline \hline
Name & Description & 11 & 12  & 13 & 14 & 15 & 16 & 17 &  \\
\hline
\multirow{2}{4em}{VGG-16}
& filter &max & fcn6 & dropout & fcn7 & dropout &fcn\_pred\\
& channel(kenel-stride) &512(2-2) & 4096(7-1) & - & 4096(1-1) & - & 19(1-1) \\
\hline
\multirow{2}{4em}{Ours}
& filter & C-dropout & fcn7\_k($\times$37) & msp1\_k($\times$34)  & msp2\_k($\times$34)   & msp3\_k($\times$34)
& elt(+)  & concat  & pred($\times19$) \\
& channel(kenel-stride)  & - & 128(1-1) & 128(7-1)  & 128(7-1) & 128(7-1) & - & - & 1(1,1)   \\
\hline
\end{tabular}
\caption{Details of our network settings and comparison with VGG-16 \cite{simonyan2014very}. 
\emph{fcn7\_k} is the filter bank for the $k^{th}$ part. 
\emph{$\times37$} represent for the $2\times 18+1$ sets of filters for two directions and the background.
\emph{msp1\_k} represents the first step of message passing layer for $k^{th}$ part.
\emph{$\times34$} are the $17\times 2$ connections on the bi-directional tree.
\emph{elt(+)} stands for element-wise summation.
This table only lists the number of filters, kernel size and stride of each layer, and the message passing process should follow Fig. \ref{fig:outline} (2,a) and (2,b).}
\label{tab. compare VGG}
\end{table*}

\subsection{Model analysis}
\subsubsection{Enlarged receptive field}
Researchers have done pose estimation at different levels: holistic (full body) \cite{toshev2014deeppose} level, poselet (combination of multiple body joints) level  \cite{Bourdev:Poslet2,Wang:PoseletsParsing,tian2012exploring,pishchulin2013poselet} and part (body joint) level \cite{yang2013articulated,felzenszwalb2005pictorial}. Latent structure \cite{tian2012exploring,wang2013beyond} and loop graph \cite{Wang:PoseletsParsing} have been employed to combine information from different scales to boost the performance. Our proposed message passing method naturally obtains features whose receptive fields are in different sizes. In this sense, it combines features at multiple scales. 

In the fcn7 layer of VGG, the receptive fields of feature maps are $188 \times 188$. When they are convolved with transform kernels, the receptive fields of the transformed features are $332 \times 332$. When the transformed features are added to the original features at a neighbor joint, features at different scales are combined. 
When features at a root joint propagate to a leaf joint through multiple convolution layer at the intermediate joints, the receptive fields get even larger. 

\subsubsection{Expressive power}
The expressive power of our transform kernels is much larger than existing message passing methods on score maps \cite{chen2014articulated}. 
Taking the settings for LSP dataset as an example, there are $128\times64\times2$ kernels between every pair of body joints while each kernel is a 7 by 7 matrix.
The message passing process also increase the depth of model. The root joint have 34 layers with multiple intermediate supervision.

\subsubsection{Relation to recurrent neural network}
Recurrent neural network (RNN) also passes information at the feature level.  It is different from ours mainly in the way of sharing weights. RNN shares feature channels at different time steps and it requires the transfer matrix between features of successive time steps to be shared among all the time steps. In our model, body joints have their own feature channels and the geometrical transform kernels are not shared. This is because feature channels for each joints have different semantic meanings and the relationships between feature maps of neighbor joints are part specific.

\section{Summary of Pipeline}
The overall pipeline is shown in Fig. \ref{fig:outline}. The ImageNet pre-trained VGG-16 \cite{simonyan2014very} is used as the base model. 
In order to keep high resolution at the prediction map, the pool4 and pool5 layers are removed from VGG. Under this setting, the feaure maps in the fcn6 layer are only downsampled by $8$ times.  Given an $448 \times 448$ input image,
the output score maps of joints are $56 \times 56$.
Channel dropout \cite{tompson2015efficient} after ReLU6 is employed to prevent overfitting. Details of the net structure are listed in Table \ref{tab. compare VGG}.

All the joints share layers up to fcn6. As shown in Fig. \ref{fig:outline} (1), in the fcn7 layer, every joint obtains its own set of $128$ feature channels on each message passing direction by convolution. These feature maps are refined through message passing in a structured feature learning layer in (2). 
The dependency of feature maps of joints is modeled with a bi-directional tree.  (2,a) and (2,b) shows the information flows along opposite directions on the tree and the process of feature update. 
Complementary features are first obtained from different flow directions separately and then combined by concatenation. 
The score map $\mathbf{z}_k$ of joint $k$ is predicted from the combined feature maps through $1 \times 1$ convolution across feature maps,
\begin{equation}
	\mathbf{z}_k = [\mathbf{A'}_k, \mathbf{B'}_k] \otimes \mathbf{w}^k_{pred}.  \label{eq:scoremap}
\end{equation}

\subsection{Details on model training}
The Conv1\_1 to fcn6 used pre-trained weights as initialization and the all the other layers are random initialized. They are finetuned together. The lower layers used pre-trained weights are finetuned with an initial learning rate of 0.001 and the newly initialized layers used an initial learning rate of 0.01.
\begin{figure}[t!]
\begin{center}
   \includegraphics[width=0.8\linewidth]{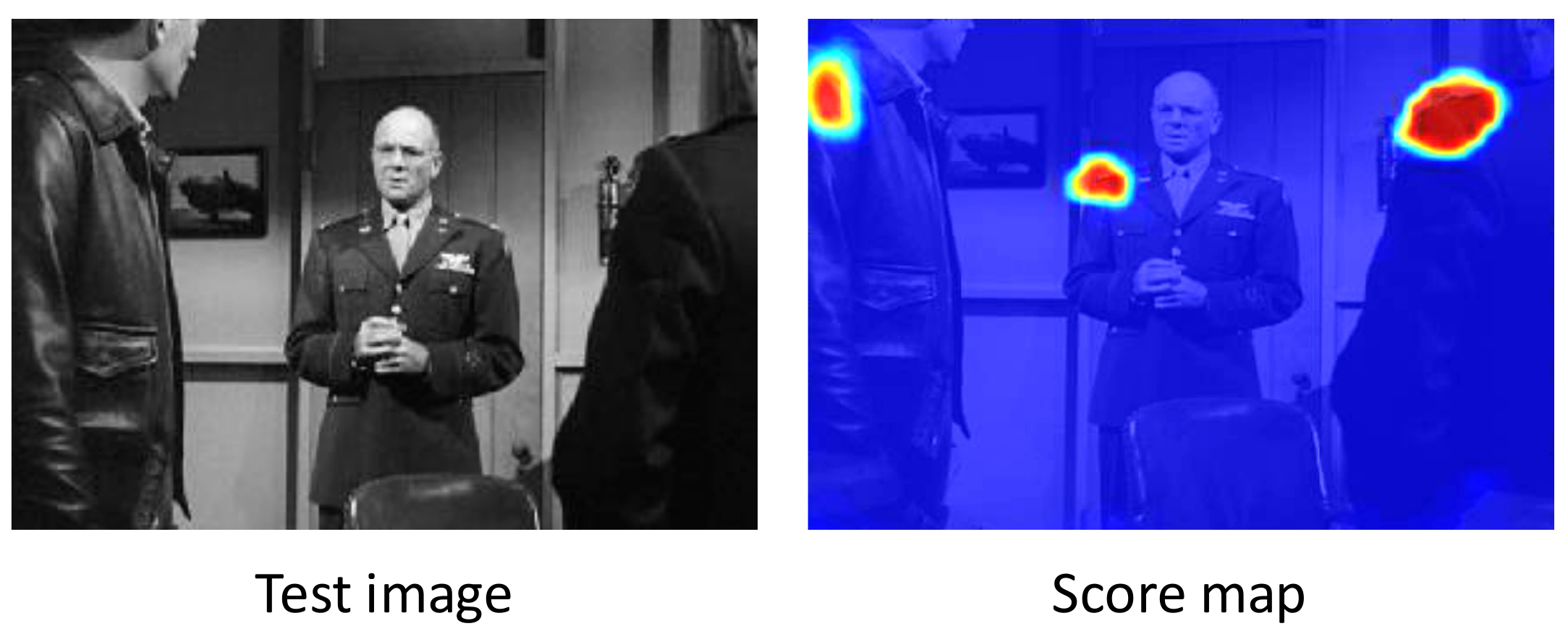}
\end{center}
\vspace{-2 mm}
   \caption{Test score map. On the left is the input image the right is the score map for shoulder on the left.}
\label{fig:test_process}
\end{figure}

\subsection{Post-processing}
A direct way of obtaining the location of a body joint is to search for the location with the maximum value on the score map for the joint. However, there is a problem when an input image has multiple persons as shown in Fig. \ref{fig:test_process}. Although the score map is clear without false alarms, it has three high response regions on three shoulders of different persons. Directly searching for maximum values on score maps separately may link body joints of different persons. It cannot be solved at the feature level and needs structural reasoning on score maps. 
It indicates that structural learning at the feature level and the score level are complementary. 
A simple post-processing is used to handle this problem. We use the distance descriptor $[(dx)^2, (dy)^2]$ to constrain the distance among body joints.  $dx=(x_i-x_j-x_r)$ and $dy=(y_i-y_j-y_r)$, where $(x_i, y_i)$ and $(x_j, y_j)$ are the locations for body joints $i$ and $j$, and $(x_r, y_r)$ is the mean relative position between body joints $i$ and $j$.
The weights for the descriptor $[dx^2, dy^2]$ are fixed as $[0.01,0.01]$.
This score map post-processing is very simple comparing with the approaches in \cite{yang2013articulated, chen2014articulated}.


\section{Training}
In order to train the network, the localization of body joints is formulated as a classification problem.
The supervision for an input image is a label tensor in size of $56 \times 56 \times 19$.
The first 18 channels represent for 18 human body joints and the $19^{th}$ channel represents for the background.
Each pixel is assigned with a class label.
The objective is to minimize the following function:
\begin{equation}
   \sum_x\sum_y m(x,y) \sum_{k} t_k(x,y) log(\frac{e^{z_k(x,y)}} {\sum_{k'}e^{z_{k'}(x,y)}})
\end{equation}
where $\{(x,y)\}$ are locations, and $k\in\{1,2,...19\}$ is the class index. $t_k(x,y)$ is the ground truth label at location $(x,y)$. 
$t_k(x,y)=1$ if $(x,y)$ belongs to class $k$, and 0 otherwise. $z_{k}(x,y)$ is the score value obtained in Eq. (\ref{eq:scoremap}). Since the number of negative training samples is far larger than the positive ones, $\mathbf{m}$ is a binary mask only keep $0.05\%$ negative samples by random selection.


\section{Experimental Results}

We show experimental results on two public human pose estimation benchmarks: the ``Frames Labeled In Cinema'' (FLIC) \cite{sapp2013modec} dataset and the ``Leeds Sports Poses'' (LSP) dataset \cite{johnson2010clustered}. 
We also provide model components analysis based on the FLIC dataset. 
On the FLIC and the LSP datasets, the Percentage of Correct Parts (PCP), the most popular evaluation criterion, is employed.
We also show results of elbow and hand using the percentage of detected body joints (PDJ) evaluation criteria on the FLIC dataset.
For the evaluation metric PCP, there are several different interpretations, which lead to a large variance in the performance.
Here, we use the \emph{strict} PCP: 
only if both ends of a limb lie within 50\% of the length of the ground-truth annotation, will this prediction be considered as correct.

\subsection{Experimental results on FLIC dataset}
\label{Sec: Flic}

\begin{figure}[t]
\begin{center}
\includegraphics[width=1\linewidth]{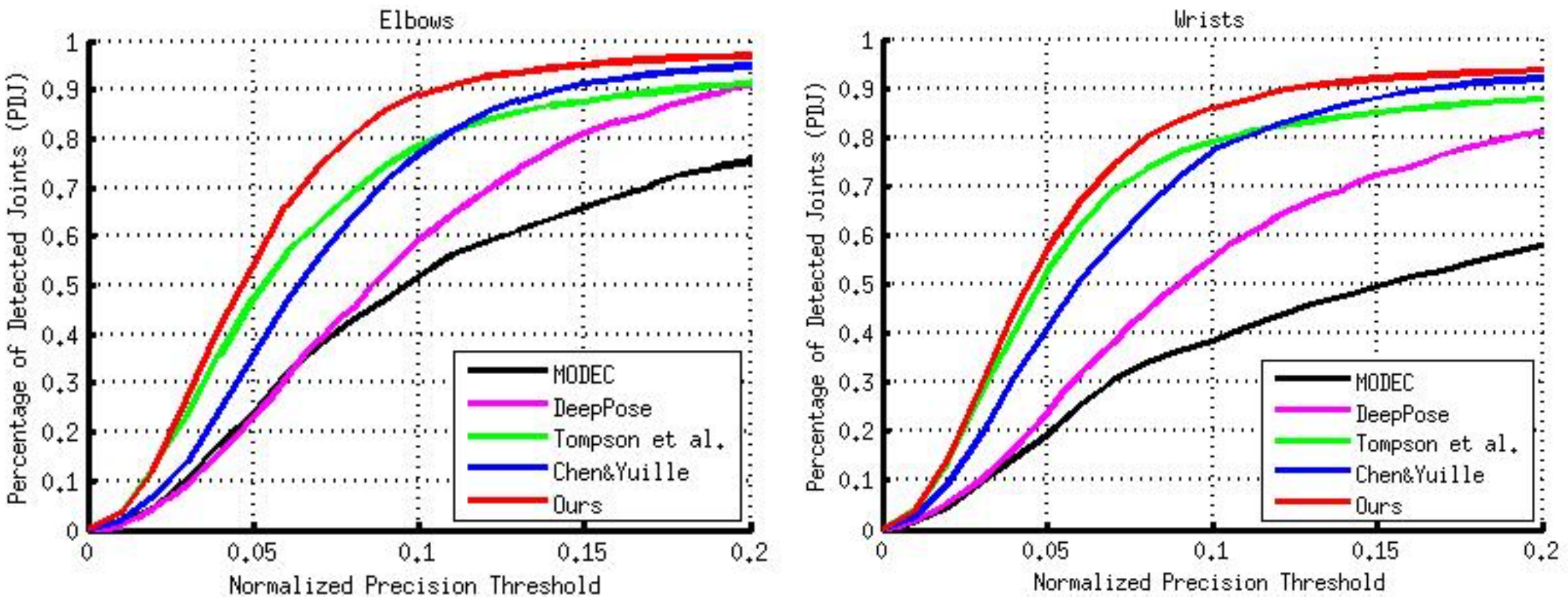}
\end{center}
\vspace{-2 mm}
\caption{Cross-data PDJ comparison of elbows and wrist on the FLIC dataset. The curves include results from MODECT \cite{sapp2013modec}, Deep pose \cite{toshev2014deeppose}, Tompson \etal \cite{tompson2014joint}, Chen\&Yuille \cite{chen2014articulated} and ours.}
\label{fig:PDJ}
\end{figure}
\begin{table}[t]\small
\centering
\begin{tabular}{l | p{0.6cm}<{\centering} | p{0.6cm}<{\centering} | p{0.8cm}<{\centering} | p{0.8cm}<{\centering} | p{0.6cm}<{\centering} }
\hline\hline
Experiment   & Head & Torso & U.arms & L.arms & Mean  \\
\hline
MODEC \cite{sapp2013modec} & - & -  & 84.4   & 52.1   & 68.3 \\
Tompson \etal \cite{tompson2014joint} &- & -	& 93.7	 & 80.9	 &87.3 \\
Tompson \etal \cite{tompson2015efficient} & - & - & 94.7 & 82.8 & 88.8\\
Chen\&Yuille \cite{chen2014articulated}  & - & - & 97.0 & 86.8 & 91.9 \\
Ours & \textbf{98.6} & \textbf{93.9} & \textbf{97.9} & \textbf{92.4} & \textbf{95.2} \\
\hline
\end{tabular}
\vspace{2 mm}
\caption{Comparison of \emph{strict} PCP results on the FLIC dataset for our method and previous approaches. Note that previous works only evaluate the performance of U.arms and L.arms, so the \emph{Mean} is the average result for U.arms and L.arms} 
\label{tab. FLIC compare with previous work}
\end{table}

The FLIC \cite{sapp2013modec} dataset contains 5002 images extracted from Hollywood movies with a person detector.
Each person is annotated with 10 body joints on the upper body and this annotation is observe-centric.
3987 images are used for training and 1016 images are used for testing. 
We augment the training images with flipping and rotation.
The INRIA \cite{histogramsINRIA} negative samples are also used in the training and validation data.
In the testing stage, the person detection results are provided for evaluation. 
We return the highest prediction with neck lying in the person detection box region, which is the same as the method in Chen and Yuille \cite{chen2014articulated}.
We linearly interpolate body joints from the 10 labeled joints to 18 joints.

\subsubsection{Overall results on FLIC}

Comparison of our method with previous works under the PCP evaluation criterion is shown in Tab. \ref{tab. FLIC compare with previous work}.
The work of Chen and Yuille \cite{chen2014articulated} and the work of Tompson \etal \cite{tompson2014joint,tompson2015efficient} are based on CNN features as well.
Tompson \etal \cite{tompson2015efficient} used 3-resolutions.
Our method performs better than all previous works and improves the performance to 95.2\%, $3.3$ points higher than the previously best approach.
It should be mentioned that any improvement gained based on $91.9$ \cite{chen2014articulated} is hard. 

We also compare our results with previous works under the PDJ evaluation criteria . 
PDJ measures the performance with a curve. The horizontal axis is the normalized precision threshold. This threshold is normalized by ground-truth pose scale to make it sample invariant.
The vertical axis is the percentage of correctly detected joints. Thus PDJ evaluates the number of body joints considered to be correct as a function of the precision threshold.
Fig. \ref{fig:PDJ} shows cross method comparison of PDJ curves for elbows and wrists.
Our method is denoted with the red line. 
It out-performs all previous methods on every normalized precision threshold.

\subsubsection{Investigation on the components in our approach}

\begin{table}[ht]\small
\centering
\begin{tabular}{l | p{1cm}<{\centering} | c | c | c | c }
\hline\hline
Experiment   & Head & Torso & U.arms & L.arms & Mean  \\
\hline
Baseline & 83.5 & 71.6  & 83.8   & 66.3   & 75.1 \\
SD      & 97.4 & 89.6 & 96.1 & 79.6 & 87.9 \\
Bi-direct & 97.7 & 93.8 & 96.8 & 90.0 & 93.4\\
Bi-direct(+) & \textbf{98.6} & \textbf{93.9} & \textbf{97.9} & \textbf{92.4} & \textbf{95.2} \\
\hline
\end{tabular}
\vspace{2 mm}
\caption{Comparison of \emph{strict} PCP results on the FLIC dataset for model components investigation. Note that the \emph{Mean} is only the average result for U.arms and L.arms} 
\label{tab. Investigation Results on FLIC}
\end{table}

\begin{figure*}[t]
\begin{center}
   \includegraphics[width=1\linewidth]{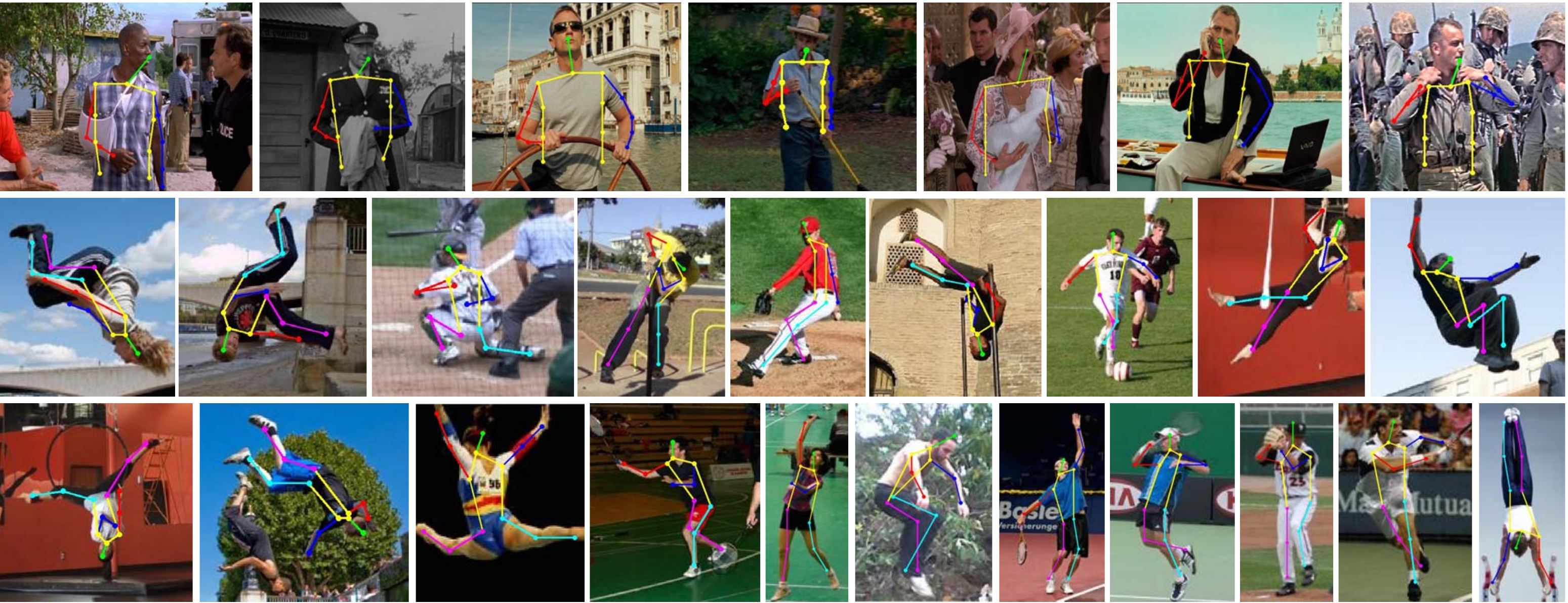}
\end{center}
\vspace{-2 mm}
   \caption{Qualitative results on FLIC and LSP datasets. The first row are results from FLIC dataset. The second and third rows are results from LSP dataset. More results can be avalibale in the supplymentary material}
\label{fig:QualitiveRes}
\end{figure*}

Model component analysis on the FLIC dataset is shown in Tab. \ref{tab. Investigation Results on FLIC}.
\emph{Baseline} is the result that directly uses the ConvNet features without structured feature learning to obtain score maps and then use our simple post processing to obtain the final result.
The result is \emph{75.1}.

The effects of structured feature learning with message passing in single upward direction are shown in Tab .\ref{tab. Investigation Results on FLIC}, denoted by \emph{SD}.
The results for this model has mean 87.9\%. 
Comparing \emph{SD} with \emph{baseline}, we observe that the PCP for each body joint in \emph{SD} is higher than \emph{VGG-baseline} and the mean PCP is improved by \textbf{13\%}.
This improvement validates the effectiveness of building up structures at the feature level.
By jointly learning structure and feature, the prediction of all body joints are better than the baseline.
The improvement comes from not only the fact that the original feature maps receive extra information from other joints for further refinement, but also that feature channels themselves are better trained when structures are modeled. 

Combination of the two directions leads to significant improvement. 
The results of bi-direct tree-structured model are denoted with \emph{Bi-direct} in Tab. \ref{tab. Investigation Results on FLIC}. 
The bi-direct model has PCP 93.4\%, \textbf{5.4\%} improvement compared with the single branch model.
Furthermore, the performance of each body joint is consistently improved compared to previous experiments.

The results discussed above use only one score map for a body joint in both training and testing. We can also  produce multiple score maps for a single body joint by clustering the body joint into appearance mixtures. We use the approach in \cite{yang2013articulated} for obtaining appearance mixtures.  For each  joint, we calculate the relative location of the current joint to its parent node, and normalize this distance with head scale. And the relative location is used for clustering each body joint into 13 mixture types with k-means.
The experimental results of model trained with multiple score maps of a body joint are shown in Tab. \ref{tab. Investigation Results on FLIC}, denoted by \emph{Bi-direct+}. The use of multiple score maps leads to 1.8\% further improvement compared with the use of single score map per joint.

\begin{table*}[t]\small
\centering
\begin{tabular}{l | p{1.3cm}<{\centering} | p{1.3cm}<{\centering} | p{1.3cm}<{\centering} | p{1.3cm}<{\centering} | p{1.3cm}<{\centering} | p{1.3cm}<{\centering} | p{1.3cm}<{\centering}}
\hline\hline
Experiment   & Torso & Head & U.arms & L.arms & U.legs & L.legs & Mean  \\
\hline
  
Andriluka \etal \cite{Andriluka:Pictorial}   &80.9 & 74.9 & 46.5 & 26.4 & 67.1 & 60.7 & 55.7 \\
Yang\&Ramanan \cite{yang2013articulated}   &82.9 & 79.3 & 56.0 & 39.8 & 70.3 & 67.0 & 62.8 \\
Pishchulin \etal \cite{pishchulin2013poselet} &87.5 &78.1 & 54.2 & 33.9 & 75.7 & 68.0 & 62.9 \\
Eichner\&Ferrari \etal \cite{eichner2012appearance} &86.2 &80.1 &56.5 &37.4 &74.3 &69.3 &64.3 \\
Ouyang \etal \cite{ouyang2014multi}      &85.8   &83.1   &63.3  & 46.6 & 76.5 & 72.2 & 68.6 \\
Pishchulin \etal \cite{pishchulin13iccv} & 88.7  & 85.1  & 61.8 & 45.0 & 78.9 & 73.2 & 69.2 \\
Chen\&Yuille \cite{chen2014articulated}  & 92.7  & 87.8  & 69.2 & 55.4 & 82.9 & 77.0  & 75.0 \\
Ours & \textbf{95.4} & \textbf{89.4} & \textbf{76.0} & \textbf{64.3} & \textbf{87.6} 
         & \textbf{83.5} & \textbf{80.8}\\
\hline
\end{tabular}
\vspace{2 mm}
\caption{Experimental results on the LSP dataset under the evaluation criteria \emph{strict} PCP } 
\label{tab. LSP}
\end{table*}

\subsection{Experimental results on LSP dataset}
LSP \cite{johnson2010clustered} is a benchmark whose images are from sport activities with full body.
It contains 2000 images, 1000 for training and 1000 for test.
Persons in this dataset are annotated with full body joints.
In the experiment, we interpolate joints on limbs and torso. Hence the total number of body joints used is 26.
In the training data, 800 images are used for training and 200 images for validation. 
Given the small amounts of samples available and large amounts of weights to be learned, we do a large amount of data augmentation.
As in \cite{chen2014articulated}, each training image is first flipped horizontally and then rotated by 360 degrees.
We also use INRIA negative images as negative samples, which were also used the existing works.
The resolution of images from the LSP dataset is smaller than FLIC, so we use a smaller size of input, i.e. $336 \times 336$, and the corresponding output score map is of size $42 \times 42$.
The images are resized to have the longer side being 336.
Given the smaller size of label map, the convolution kernel size is also changed. 
Each geometrical transform is implemented with two steps of convolutions with kernel size $7 \times 7$ on LSP. 

PCP results are shown in Tab. \ref{tab. LSP}. 
The work of Chen and Yuille \cite{chen2014articulated} also used the deep model. The other works were based on hand-crafted features.
We do not compare with DeepPose \cite{toshev2014deeppose} because their work used person-centric training and evaluation, while all the works mentioned in Tab. \ref{tab. LSP} including ours are observe-centric.
Our method outperforms previous state-of-the-art by 5.8\%.
It also obtains the best result on every body part evaluated.

\section{Conclusion}
We propose the idea of modeling correlations among feature maps of body joints for pose estimation. 
Feature level information passing delivers more detailed descriptions about body joints than score maps.
It is implemented with geometrical transform kernels. 
A bi-directional tree structured model is proposed and complementary features are learned from information flow in opposite directions. 
Experimental results on two public datasets show that the proposed framework improves feature learning substantially. Even with very simple post processing, it outperforms the state-of-the-art method. 
In the future work, further improvement is expected by integrating with more advanced post processing techniques from existing literature. Moreover, various structures for message passing investigated in existing works \cite{yang2013articulated,chen2014articulated,pishchulin2013poselet} could also provide guidance to improve message passing at the feature level.  

\textbf{Acknowledgment}:
This work is partially support by SenseTime Group Limited and the General Research Fund sponsored by the Research Grants Council of Hong Kong (Project Nos. CUHK14206114, CUHK14205615, CUHK417011, CUHK14207814, CUHK14203015). 



{\small
\clearpage
\bibliographystyle{ieee}
\bibliography{egbib}
}

\end{document}